\documentclass{article}

\pdfoutput=1

\usepackage[ruled]{algorithm}
\usepackage[noend]{algorithmic}
\usepackage[pdftex]{graphicx}
\usepackage{amsmath,amssymb,amstext,amsthm}
\usepackage[utf8]{inputenc}
\usepackage{multirow}
\usepackage{verbatim}
\usepackage[margin=3cm]{geometry}

\newcommand{\nt}[1]{\mbox{{\tt [#1]}}}
\newcommand{\Right}{\ensuremath{\rightarrow} \,}
\newcommand{\Upright}{\ensuremath{\nearrow}  \,}
\newcommand{\Downright}{\ensuremath{\searrow} \,}
\newcommand{\Down}{\ensuremath{\downarrow} \,}
\newcommand{\Contains}{\ensuremath{\odot} \,}

\DeclareMathOperator{\overlap}{overlap}
\DeclareMathOperator{\score}{sc}
\DeclareMathOperator{\class}{c\ell}

\newcommand{\produces}[1]{\ensuremath{\overset{#1}{\Rightarrow}}}

\renewcommand{\Pr}[1]{\ensuremath{P \left( #1 \right)}}
\newcommand{\CondPr}[2]{\ensuremath{P \left( #1 \mid #2 \right)}}
\newcommand{\nil}{\textsc{nil}}
\newcommand{\expr}{\textsc{expr}}
\newcommand{\sym}{\textsc{sym}}
\newcommand{\gen}{\textsc{gen}}

\newtheorem{definition}{Definition}

\title{A Bayesian model for recognizing handwritten mathematical expressions}
\author{Scott MacLean \and George Labahn}
\date{Cheriton School of Computer Science \\ University of Waterloo \\ 200 University Ave. W. \\ Waterloo, ON, Canada\\N2L 3G1\\\{smaclean,glabahn\}@uwaterloo.ca}

\begin{document}

\maketitle
\begin{abstract}
Recognizing handwritten mathematics is a challenging classification
problem, requiring simultaneous identification of all the symbols comprising an
input as well as the complex two-dimensional relationships between symbols and
subexpressions. Because of the ambiguity present in handwritten input, it is
often unrealistic to hope for consistently perfect recognition accuracy. We
present a system which captures all recognizable interpretations of the input
and organizes them in a parse forest from which individual parse trees may be
extracted and reported. If the top-ranked interpretation is incorrect, the user
may request alternates and select the recognition result they desire. The tree
extraction step uses a novel probabilistic tree scoring strategy in which a
Bayesian network is constructed based on the structure of the input, and each
joint variable assignment corresponds to a different parse tree. Parse trees
are then reported in order of decreasing probability. Two accuracy evaluations
demonstrate that the resulting recognition system is more accurate than previous
versions (which used non-probabilistic methods) and other academic math recognizers.
\end{abstract}


\section{Introduction}
\label{sec:introduction}

Many software packages exist which operate on mathematical expressions. Such
software is generally produced for one of two purposes: either to create
two-dimensional renderings of mathematical expressions for printing or
on-screen display (e.g., \LaTeX, MathML), or to perform mathematical operations on the
expressions (e.g., Maple, Mathematica, Sage, and various numeric and symbolic
calculators). In both cases, the mathematical expressions themselves must be
entered by the user in a linearized, textual format specific to each software package.

This method of inputting math expressions is unsatisfactory for two main
reasons. First, it requires users to learn a different syntax for each software
package they use. Second, the linearized text formats obscure the
two-dimensional structure that is present in the typical way users draw math
expressions on paper. This is demonstrated with some absurdity by software
designed to render math expressions, for which one must linearize a two-dimensional
expression and input it as a text string, only to have the software re-create and
display the expression's original two-dimensional structure.

Mathematical software is typically used as a means to accomplish a particular
goal that the user has in mind, whether it is creating a web site with
mathematical content, calculating a sum, or integrating a complex expression.
The requirement to input math expressions in a unique, unnatural format is
therefore an obstacle that must be overcome, not something learned for its own
sake. As such, it is desirable for users to input mathematics by drawing
expressions in two dimensions as they do with pen and paper.

Academic interest in the problem of recognizing hand-drawn math expressions
originated with Anderson's doctoral research in the late 1960's \cite{anderson68}.
Interest has waxed and waned in the intervening decades, and recent years have
witnessed renewed attention to the topic, potentially spurred on by the
nascent Competition on Recognition of Handwritten Mathematical Expressions
(CROHME) \cite{crohme2011,crohme2012}.

However, math expressions have proved difficult to recognize effectively.
Even the best state of the art systems (as measured by CROHME) are not sufficiently
accurate for everyday use by non-specialists. The recognition problem is
complex as it requires not only symbols to be recognized, but also
arrangements of symbols, and the semantic content that those arrangements
represent. Matters are further complicated by the large symbol set and the
ambiguous nature of both handwritten input and mathematical syntax.

\subsection{MathBrush}
\label{sec:mathbrush}

\begin{figure}[htp]
\centering
\includegraphics[width=0.6\linewidth]{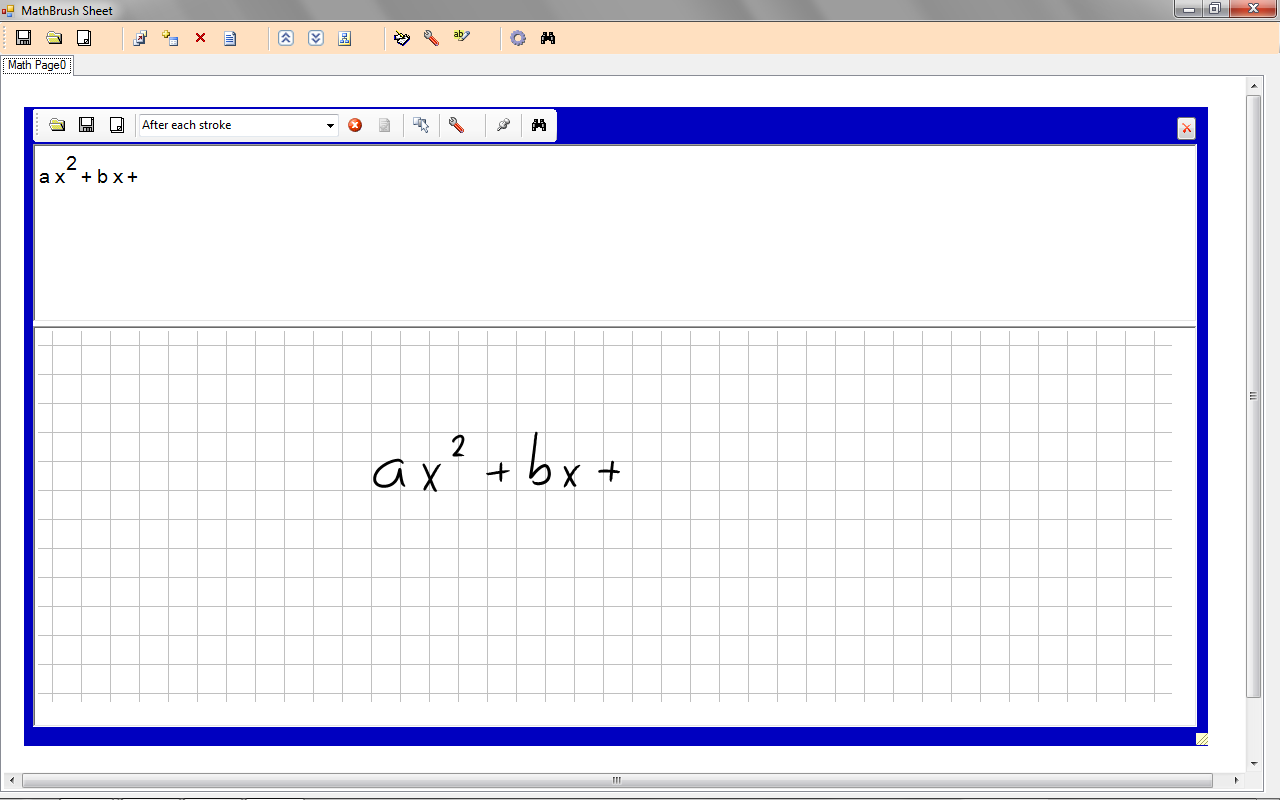} \\
\includegraphics[width=0.6\linewidth]{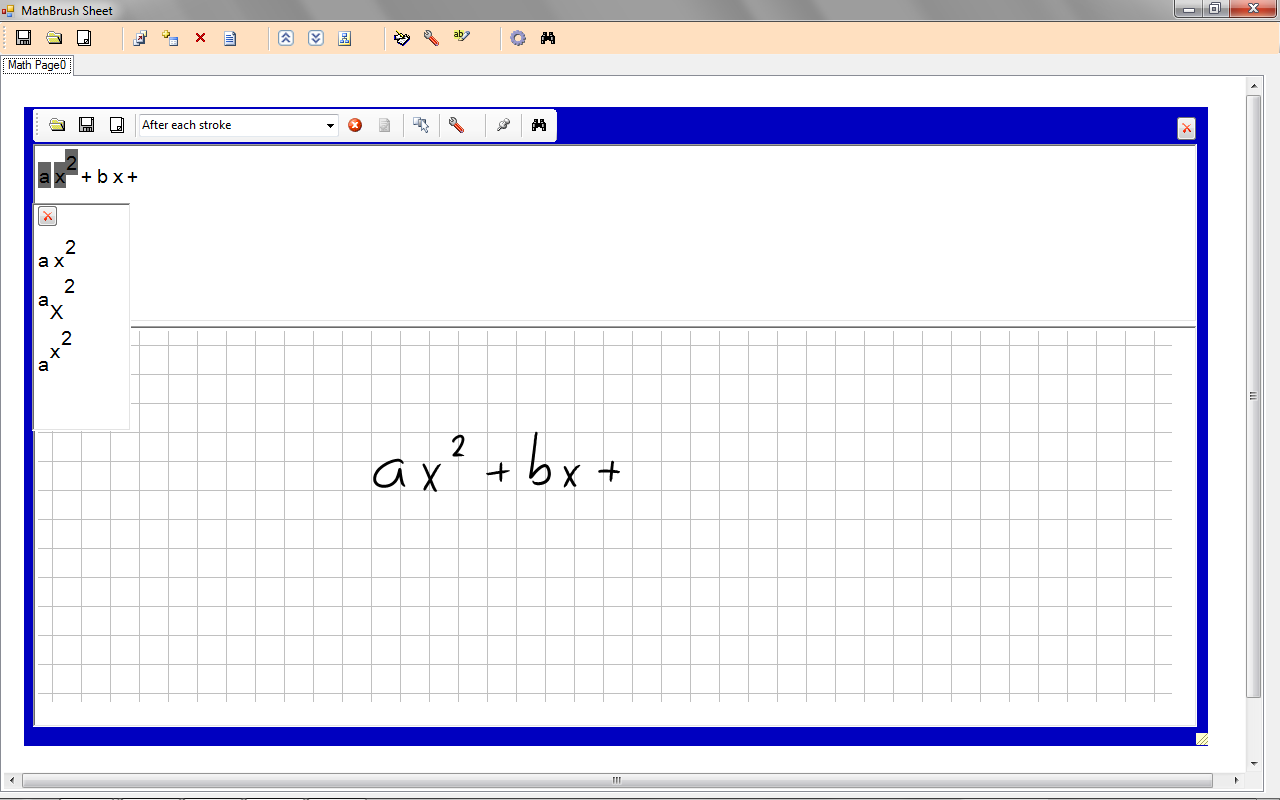} \\
\includegraphics[width=0.6\linewidth]{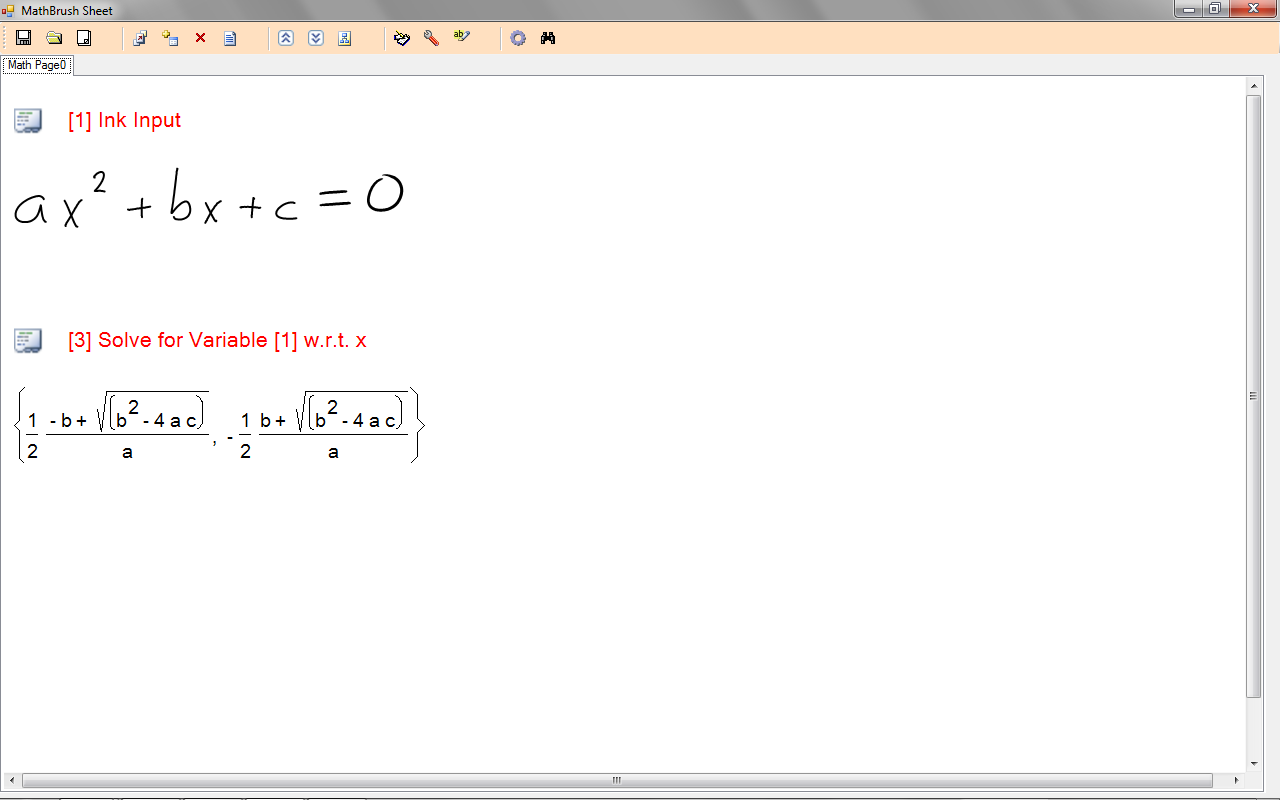}
\caption{\label{fig:mathbrush}MathBrush. From top to bottom: (a) input interface, (b) correction drop-down, (c)
worksheet interface.}
\end{figure}

Our work is done in the context of the MathBrush pen-based math system
\cite{labahn08}. Using MathBrush, one draws an expression, which is recognized incrementally as it is
drawn (Fig. \ref{fig:mathbrush}a). At any point, the user may correct erroneous
recognition results by selecting part or all of the expression and choosing an
alternative recognition from a drop-down list (Fig. \ref{fig:mathbrush}b). Once the expression is completely
recognized, it is embedded into a worksheet interface through which the user
may manipulate and work with the expression via computer algebra system
commands (Fig. \ref{fig:mathbrush}c).

For our purposes, the crucial element of this description is that the user may
at any point request alternative interpretations of any portion of their
input. This behaviour differs from that of most other recognition systems,
which present the user with a single, potentially incorrect result. We instead
acknowledge the fact that recognition will not be perfect all of the time, and
endeavour to make the correction process as fast and simple as possible.
Much of our work is motivated by the necessity of maintaining multiple
interpretations of the input so that lists of alternative recognition
results can be populated if the need arises.

We have previously reported techniques for capturing and reporting these multiple
interpretations \cite{ijdar2013}, and some of those ideas remain the foundation
of the work presented in this paper.
In particular, the two-dimensional syntax of mathematics is organized
and understood through the formalism of relational context-free grammars
(RCFGs), a multi-dimensional generalization of traditional context-free
grammars. It is necessary to place restrictions on
the structure of the input so that parsing with RCFGs takes a reasonable
amount of time, but once this is done it is fairly straightforward to adapt
existing parsing algorithms to RCFGs so that they produce a data structure called a
parse forest which simultaneously represents all recognizable interpretations
of an input. The parse forest is the primary tool we use to organize
alternative recognition results in case the user desires to see them. When it
does become necessary to populate drop-down lists in MathBrush with
recognition results, individual parse trees -- each representing a different
interpretation of the input -- must be extracted from the parse forest. As with
parsing, some restrictions are necessary in this step to achieve reasonable
performance.

Our previous work formalized the notion of ambiguity in terms of fuzzy sets
and represented portions of the input as fuzzy sets of math expressions. In this
paper, we abandon the language of fuzzy sets, develop a more general
notion of how recognition processes interact with the grammar model, and
introduce a Bayesian scoring model for parse trees. The scoring model
incorporates results from recognition systems as well as ``common-sense''
knowledge drawn from training corpora. The recognition systems
contributing to the model include a novel stroke grouping system, a combination
symbol classifier based on quantile-mapped distance functions, and a naive Bayesian
relation classifier. An algebraic technique eliminates many model
variables from tree probability calculations, yielding an efficiently
computable probability function. The resulting recognition
system is significantly more accurate than other academic math recognizers,
including the fuzzy variant we used previously.

The rest of this paper is organized as follows. Section
\ref{sec:existing-research}
summarizes some recent probabilistic approaches to the math recognition
problem and points out some of the issues that arise in practice when applying standard
probabilistic models to this problem. Section \ref{sec:grammars-and-parsing}
summarizes the theory and algorithms behind our RCFG variant, describing how
to construct and extract trees from a parse forest for a given input. This
section also states the restrictions and assumptions we use to attain
reasonable recognition speeds.
Section \ref{sec:scoring-function} is devoted to the new probabilistic tree scoring
function we have devised for MathBrush, based on constructing a Bayesian network
for a given input.
Finally, Section \ref{sec:evaluation} replicates the 2011 and 2012 CROHME
accuracy evaluations, demonstrating the effectiveness of the probabilistic
scoring framework, and Section \ref{sec:conclusions} concludes the paper with
a summary and some promising directions for future research.


\section{Existing research}
\label{sec:existing-research}

There is a significant and expanding literature concerning the math
recognition problem. Many recent systems and approaches are summarized in the
CROHME reports \cite{crohme2011,crohme2012}. We will comment directly on a few
recent approaches using probabilistic methods.

\subsection{Alvaro et al}

The system developed by Alvaro et al \cite{alvaro11} placed first in the CROHME 2011
recognition contest \cite{crohme2011}. It is based on earlier work of Yamamoto et al
\cite{yamamoto06}. A grammar models the formal structure of math
expressions. Symbols and the relations between them are modeled stochastically
using manually-defined probability functions. The symbol recognition
probabilities are used to seed a parsing table on which a CYK-style parsing algorithm
proceeds to obtain an expression tree representing the entire input.

In this scheme, writing is considered to be a generative stochastic process
governed by the grammar rules and probability distributions. That is, one
stochastically generates a bounding box for the entire expression, chooses a
grammar rule to apply, and stochastically generates bounding boxes for each of
the rule's RHS elements according to the relation distribution. This process
continues recursively until a grammar rule producing a terminal symbol is
selected, at which point a stroke (or, more properly, a collection of stroke
features) is stochastically generated.

Given a particular set of input strokes, Alvaro et al find the 
sequence of stochastic choices most likely to have generated the input.
However, stochastic grammars are known to biased toward short
parse trees (those containing few derivation steps) \cite{manning99}. In our
own experiments with such approaches, we encountered difficulties in
particular with recognizing multi-stroke symbols in the context of full
expressions. The model has no intrinsic notion of symbol segmentation, and
the bias toward short parse trees caused the recognizer to consistently report
symbols with many strokes even when they had poor recognition scores. Yet to introduce
symbol segmentation scores in a straightforward way causes probability
distributions to no longer sum to one. Alvaro et al allude to similar
difficulties when they mention that their symbol recognition probabilities
had to be rescaled to account for multi-stroke symbols.

In Section \ref{sec:scoring-function}, we propose a different solution. We
abandon the generative model in favour of one that explicitly reflects the
stroke-based nature of the input, and furthermore include a $\nil$ value to
account for groups of strokes that are not, in fact, symbols.

\subsection{Awal et al}

While the system described by Awal et al \cite{awal10} was included in the CROHME
2011 contest, its developers were directly associated with the contest and
were thus not official participants. However, their system scored higher than
the winning system of Alvaro et al, so it is worthwhile to examine its
construction.

A dynamic programming algorithm first proposes likely groupings
of strokes into symbols, although it is not clear what cost function the
dynamic program is minimizing. Each of the symbol groups is recognized using
neural networks whose outputs are converted into a probability distribution
over symbol classes.

Math expression structure is modeled by a context-free grammar in which each
rule is linear in either the horizontal of vertical direction. Spatial
relationships between symbols and subexpressions are modeled as two independent
Gaussians on position and size difference between subexpression bounding boxes.
These probabilities along with those from symbol recognition are treated as
independent variables, and the parse tree minimizing a cost function defined in
terms of weighted negative log likelihoods is reported as the final parse.

This method as a whole is not probabilistic as the
variables are not combined in a coherent model. Instead, probability distributions
are used as components of a scoring function. This pattern is common in the
math recognition literature: distribution functions are used when they are
useful, but the overall strategy remains ad hoc. Vestiges of this may be seen
in our own work throughout Section \ref{sec:scoring-function}; however in this
paper we take the next step and re-combine scoring functions into a coherent
probabilistic model.

\subsection{Shi, Li, and Soong}

Working with Microsoft Research Asia, Shi, Li, and Soong proposed a unified
HMM-based method for recognizing math expressions \cite{shi07}. Treating
the input strokes as a temporally-ordered sequence, they use dynamic
programming to determine the most likely points at which to split the sequence
into distinct symbol groups, the most likely symbols each of those groups
represent, and the most likely spatial relation between temporally-adjacent
symbols. Some local context is taken into account by treating symbol and relation
sequences as Markov chains. This process results in a DAG, which may be easily
converted to an expression tree.

To compute symbol likelihoods, a grid-based method measuring point density and
stroke direction is used to obtain a feature vector. These vectors are assumed
to be generated by a mixture of Gaussians with one component for each known
symbol type. Relation likelihoods are also treated as Gaussian mixtures of
extracted bounding-box features. Group likelihoods are computed by
manually-defined probability functions.

This approach is elegant in its unity of symbol, relation, and expression
recognition. The reduction of the input to a linear sequence of strokes
dramatically simplifies the parsing problem. But this assumption of linearity
comes at the expense of generality. The HMM structure strictly requires strokes
to be drawn in a pre-determined linear sequence. That is, the model accounts
for ambiguity in symbol and relation identities, but not for the two-dimensional
structure of mathematics. As such, the method is unsuitable for applications
such as MathBrush, in which the user may draw, erase, move, or otherwise edit
any part of their input at any time.

\section{Relational grammars and parsing}
\label{sec:grammars-and-parsing}

Relational grammars are the primary means by which the MathBrush recognizer
associates mathematical semantics with portions of the input. We previously
described a fuzzy relational grammar formalism which explicitly modeled
recognition as a process by which an observed, ambiguous input is interpreted
as a certain, structured expression \cite{ijdar2013}. In this work, we define
the grammar slightly more abstractly so that it has no intrinsic tie to fuzzy
sets and may be used with a variety of scoring functions.
Interested readers are referred to \cite{ijdar2013} and \cite{phd-thesis} for
more details.

\subsection{Grammar definition and terminology}
\label{sec:grammar-definition}

Relational context-free grammars are defined as follows.

\begin{definition}
A \textit{relational context-free grammar} $G$ is a tuple $\left( \Sigma, N,
S, O, R, P \right)$, where
\begin{itemize}
	\item{$\Sigma$ is a set of terminal symbols,}
	\item{$N$ is a set of non-terminal symbols,}
	\item{$S \in N$ is the \textit{start symbol},}
	\item{$O$ is a set of observables,}
	\item{$R$ is a set of relations on $I$, the set of interpretations of $G$
	(described below),}
	\item{$P$ is a set of productions, each of the form $A_0 \produces{r} A_1
	A_2 \cdots A_k$, where $A_0 \in N, r \in R$, and $A_1,\ldots,A_k \in N \cup
	\Sigma$,}
\end{itemize}
\label{def:grammars-definition}
\end{definition}

The basic elements of a CFG are present in this definition as $\Sigma,N,S,$
and $P$, though the form of an RCFG production is slightly more complex than
that of a CFG production. However, there are several less familiar components
as well.

\subsubsection{Observables, Expressions and interpretations}
\label{sec:expressions}

The set $O$ of \textit{observables} is the set of all possible
inputs -- in our case each observable is a set of ink strokes, and
each ink stroke is an ordered sequence of points in $\mathbb{R}^2$.

An \textit{expression} is the generalization of a string to the RCFG case.
The form of an expression reflects the relational form of the grammar
productions. Any terminal symbol $\alpha \in \Sigma$ is a terminal expression.
An expression $e$ may also be formed by
concatenating several expressions $e_1,\ldots,e_k$ by a relation
$r \in R$. Such an \textit{$r$-concatenation} is written
$e_1 r \cdots r e_k$.

The \textit{representable set of $G$}, written $\mathcal{L}(G)$, is the set of all
expressions formally derivable using the nonterminal, terminals, and
productions of a grammar $G$. $\mathcal{L}(G)$ generalizes the notion of the
language generated by a grammar to RCFGs.

An \textit{interpretation}, then, is a pair $(e,o)$, where $e \in
\mathcal{L}(G)$ is a
representable expression and $o$ is an observable.
This definition links the formal structure of the
grammar with the geometric structure of the input -- an
interpretation is essentially a parse of $o$ where the structure of $e$
encodes the hierarchical structure of the parse.
The set of interpretations $I$ referenced in the RCFG definition is just
the set of all possible interpretations:
$$
I = \left\{ \left(e,o\right) : e \in \mathcal{L}(G), o \in O \right\}.
$$

In our previous work with fuzzy sets, the set of interpretations was fuzzy.
This required a scoring model to be embedded into the grammar's structure and
added otherwise unnecessary components to the grammar definition.
The present formalism is slightly more abstract. It loses no expressive power
but leaves scoring models and data organization as implementation details,
rather than as fundamental grammar components.

\subsubsection{Relations and productions}
\label{sec:relations}

\begin{figure}[htp]
\centering
\includegraphics[width=0.2\linewidth]{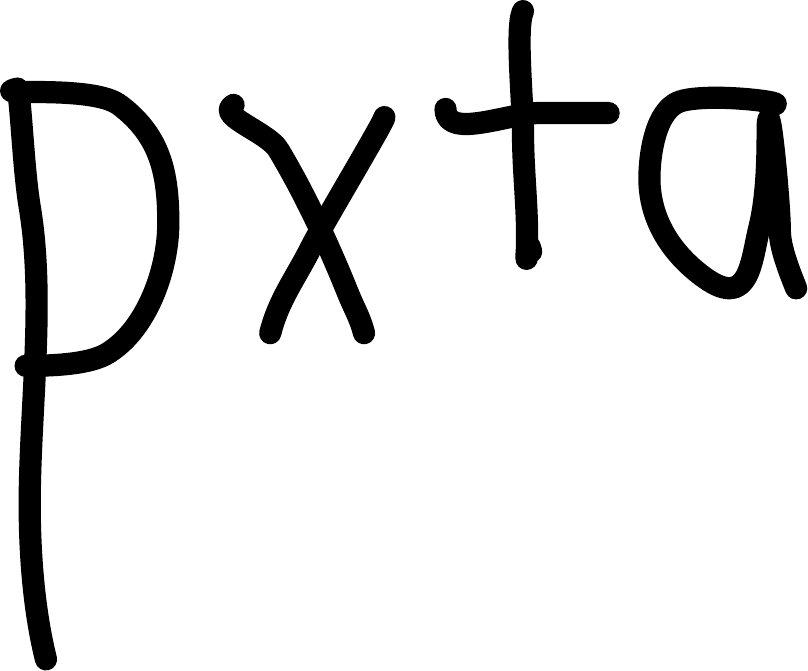}
\caption{\label{fig:grammars-context-dependence}An expression in which the optimal
relation depends on symbol identity.}
\end{figure}

The relations in $R$ model the spatial relationships between subexpressions.
In mathematics, these relationships often determine the semantic
interpretation of an expression (e.g., $a$ and $x$ written side-by-side as
$ax$ means something quite different from the diagonal arrangement
$a^x$). An important 
feature of these grammar relations is that they act on interpretations -- pairs of
expressions and observables. This enables relation classifiers to use
knowledge about how particular expressions or symbols fit together when
interpreting geometric features of observables. By doing so, an
expression like the one shown in Figure \ref{fig:grammars-context-dependence} may
be interpreted as either $P^{x+a}$ or $px+a$ depending on the identity of its
symbols.

We use five relations, denoted $\Right, \Upright, \Downright, \Down, \Contains$. The
four arrows indicate the general writing direction between two subexpressions,
and $\Contains$ indicates containment notation, as used in square roots.

The productions in $P$ are similar to context-free grammar productions. The
relation $r$ appearing above the production symbol ($\produces{r}$)
indicates that $r$ must be satisfied by adjacent
elements of the RHS. Formally, given a production $A_0 \produces{r} A_1 A_2
\cdots A_k$, if $o_i$ ($i=1,\ldots,k$) denotes an observable interpretable
as an expression $e_i$, and each $e_i$ is itself derivable from a nonterminal $A_i$,
then for the observable $o_1 \cup \cdots \cup o_k$ to be interpretable as the
$r$-concatenation $e_1 r \cdots r e_k$ requires that
$\left( \left( e_i,o_i \right), \left( e_{i+1}, o_{i+1} \right) \right) \in
r$ for $i=1,\ldots,k-1$. That is, all adjacent pairs of sub-interpretations must
satisfy the grammar relation for them to be combinable into a larger
interpretation.

For example, consider the following toy grammar:

\begin{align*}
\nt{EXPR} & \produces{} \nt{ADD} \mid \nt{TERM} \\
\nt{ADD} & \produces{\Right} \nt{TERM} + \nt{EXPR} \\
\nt{TERM} & \produces{} \nt{MULT} \mid \nt{LEAD-TERM} \\
\nt{LEAD-TERM} & \produces{} \nt{SUP} \mid \nt{FRAC} \mid \nt{SYM} \\
\nt{MULT} & \produces{\Right} \nt{LEAD-TERM} \nt{TERM} \\
\nt{FRAC} & \produces{\Down} \nt{EXPR} \text{---} \nt{EXPR} \\
\nt{SUP} & \produces{\Upright} \nt{SYM} \nt{EXPR} \\
\nt{SYM} & \produces{} \nt{VAR} \mid \nt{NUM} \\
\nt{VAR} & \produces{} a \mid b \mid \cdots \mid z \\
\nt{NUM} & \produces{} 0 \mid 1 \mid \cdots \mid 9
\end{align*}

In this example, the production for $\nt{ADD}$ models the syntax for infix
addition: two expressions joined by the addition symbol, written from left to
right. The production for $\nt{FRAC}$ has a similar structure -- two
expressions joined by the fraction bar symbol -- but specifies that its
components should be organized from top to bottom using the $\Down$ relation.
The production for $\nt{MULT}$ illustrates the ease with which different
relations and writing directions may be combined together to form larger
expressions. The $\nt{LEAD-TERM}$ term may be a superscript (using the
$\Upright$ relation), a fraction (using the $\Down$ relation), or a symbol,
but no matter which of those subexpressions appears as the leading term, it
must satisfy the left-to-right relation $\Right$ when combined with the
training term.

\subsection{Parsing with Unger's method}
\label{sec:parsing}

Compared with traditional CFGs, there are two main sources of ambiguity and
difficulty when parsing RCFGs: the symbols in the input are unknown, both in
the sense of what parts of the input constitute a symbol, and what the
identity of each symbol is; and because the input is two-dimensional, there is
no straightforward linear order in which to traverse the input strokes.
Additionally, there may be ambiguity arising from relation classification
uncertainty: we may not know for certain which grammar relation links two
subexpressions together.

We manage and organize these sources of uncertainty by using a data structure
called a parse forest to simultaneously represent all recognizable parses of the
input. A parse forest is a graphical structure in which each node represents
all of the parses of a particular subset of the input using a particular
grammar nonterminal or production. There are two types of nodes:
\begin{enumerate}
	\item{OR nodes. Each of an OR node's parse trees is taken directly from
	one of the node's children. There are two types of OR nodes:
	\begin{enumerate}
		\item{Nonterminal nodes of the form $(A,o)$, representing the parses of a
		nonterminal $A$ on a subset $o$ of the input. The parse tree of
		$A$ on $o$ is just the parse tree of some production with LHS $A$ on $o$.}
		\item{Production nodes of the form $(p,o)$, representing the parses of a
		production $p$ of the form $A_0 \produces{r} A_1 \cdots A_k$ on a subset
		$o$ of the input. Each such parse tree arises by parsing $p$ on some
		partition of $o$ into $k$ subsets.}
	\end{enumerate}
	}
	\item{AND nodes. Each of an AND node's parse trees is an $r$-concatenation
	to which each of the node's children contributes one subexpression.
	AND nodes have the form $(p;(o_1,\ldots,o_k))$, representing the parses of
	a production $p$ of the form above on a partition
	$o_1,\ldots,o_k$ of some subset $o_1 \cup \cdots \cup o_k$ of the input.
	Each subexpression $e_i$ is itself a parse of $A_i$ on the partition subset $o_i$.}
\end{enumerate}

Parsing an RCFG may be divided into two steps: forest construction, in
which a shared parse forest is created that represents all recognizable parses
of the input, and tree extraction, in which individual parse trees are
extracted from the forest in decreasing order of tree score.

\subsubsection{Restricting feasible partitions}
\label{sec:restricting-partitions}

To obtain reasonable performance when parsing with RCFGs, it is necessary to
restrict the ways in which input observables may be partitioned and
recombined (otherwise, partitions of the input using all $2^n$ subsets would
need to be considered). We choose to restrict partitions to horizontal and
vertical concatenations, as follows.

Each of the five grammar relations is associated with either the $x$ direction
of $y$ direction ($\Right,\Upright,\Contains$ with $x$, $\Downright,\Down$
with $y$). Then when parsing an observable $o$ using a production $A_0
\produces{r} A_1 \cdots A_k$, $o$ is partitioned by either horizontal or
vertical cuts based on the directionality of the relation $r$.
For example, when parsing the production $\nt{ADD} \produces{\Right}
\nt{TERM} + \nt{EXPR}$ on the input from Figure
\ref{fig:grammars-context-dependence}, we would note that the relation
$\Right$ requires horizontal concatenations, and so split the input into three
horizontally-concatenated subsets using two vertical cuts. In general, there
are $\binom{n}{k-1}$ ways to partition $n$ input strokes into $k$ concatenated subsets.

The restriction to horizontal and vertical concatenations is quite severe, but
it reduces the worst-case number of subsets of the input which may need to be
considered during parsing from $2^n$ to $\mathcal{O}(n^4)$.
Furthermore, the restriction effectively linearizes the input in either
the $x$ or $y$ dimension, depending on which grammar relation is being
considered. This linearity is sufficiently similar to the implicit linearity
of traditional CFG parsing that one may adapt CFG parsing algorithms to the
relational grammar case without much difficulty.

\subsubsection{Unger's method}
\label{sec:ungers-method}

Unger's method is a fairly brute-force algorithm for parsing CFGs \cite{unger68}.
It is easily extended to become a tabular RCFG parser as follows.

First, the grammar is re-written so that each production is either terminal
(of the form $A_0 \produces{} \alpha$, where $\alpha \in \Sigma$), or
nonterminal (of the form $A_0 \produces{r} A_1 \cdots A_k$, where each $A_i
\in N$), so there is no mixing of terminal and non-terminal symbols within
productions. Then our variant of Unger's method constructs a parse forest
by recursively applying the following rules, where $p$ is a production and $o$
is an input observable:
\begin{enumerate}
	\item{If $p$ is a terminal production, $A_0 \produces{} \alpha$, then check
	if $o$ is recognizable as $\alpha$ according to a symbol recognizer. If it
	is, then add the interpretation $(o,\alpha)$ to table entry $(o,\alpha)$;
	otherwise parsing fails.}
	\item{Otherwise, $p$ is of the form $A_0 \produces{r} A_1 \cdots A_k$.
	For every relevant partition $o$ into subsets $o_1,\ldots,o_k$ (either a
	horizontal or vertical concatenation, as appropriate), parse
	each nonterminal $A_i$ on $o_i$. If all of the sub-parses succeed, then
	add $\left(p;(o_1,\ldots,o_k)\right)$ to table entry $(o,A_0)$.
	\label{itm:parsing-recursive-parse}
	}
\end{enumerate}

Naively, $\mathcal{O}(|o|^{k-1})$ partitions must be considered in step 2.
But since Unger's method explicitly enumerates these partitions, we may
easily optimize the algorithm by restricting which partitions are explored,
thereby avoiding the recursive part of step 2 when possible. In bottom-up
algorithms like CYK or Earley's method, such optimizations are not possible
(or are at least much more difficult to implement) because the subdivision of
the input into subexpressions is determined implicitly from the bottom up.

The output of this variant of Unger's method is a parse forest containing
every parse of the input which could be recognized. The next step is to
extract individual parse trees from the parse forest in decreasing score
order. The first, highest-scoring tree is presented to the user as they are
writing (see Figure \ref{fig:mathbrush}a). If the user selects a portion of
the expression and requests alternative recognition results, then further
parse trees are extracted and displayed in the drop-down lists.

\subsection{Parse tree extraction}
\label{sec:tree-extraction}

To find a parse tree of the entire input observable $o$, we need only start at
the root node $(S,o)$ of the parse forest ($S$ being the grammar's start symbol)
and, whenever we are at an OR node, follow the link to one of the node's children.
Whenever we are at an AND node, follow the links to all of the node's children
simultaneously. Once all branches of such a path reach terminal expressions, we
have found a parse tree. Moreover, any combination of choices at OR nodes that
yields such a branching path gives a valid parse tree.

The problem, then, is how to enumerate those paths in such a way that the most
reasonable parse is found first, then the next most reasonable, and so on. 
Our approach is to equip each of the nodes in the parse forest with a priority queue
ordered by parse tree score. Tree extraction is then treated as an iteration
problem. Initially, we insert the highest-scoring parse tree of a node
into the node's priority queue. Then, whenever a tree is required from that
node, it is obtained by popping the highest-scoring tree from the queue. The
queue is then prepared for subsequent iteration by pushing in some more trees
in such a way that the next-highest-scoring tree is guaranteed to be present.

More specifically, to find the highest-scoring tree at an OR node $N$, we recursively find
the highest-scoring trees at all of the node's children and add them to the queue. Then
the highest-scoring tree at $N$ is just the highest-scoring tree out of those
options. When a tree is popped from the priority queue, it was obtained from a
child node of $N$ (call it $C$) as, say, the $i$th highest-scoring tree at
$C$. To prepare for the next iteration step, we obtain the $i+1$st highest-scoring tree at
$C$ and add that into the priority queue of $N$.

To find the highest-scoring tree at an AND node $N$ representing parses of
$A_0 \produces{r} A_1 \cdots A_k$ on a partition $o_1,\ldots,o_k$, we again
recursively find the highest-scoring parses of the node's children -- in this
case the OR nodes $(A_1,o_1),\ldots,(A_k,o_k)$. Calling these expressions
$e_1,\ldots,e_k$, the $r$-concatenation $e_1 r \cdots r e_k$ is the
highest-scoring tree at $N$. Now, any $r$-concatenation obtained from the $k$
child nodes of $N$ can be represented by a $k$-tuple $(n_1,\ldots,n_k)$,
where each $n_i$ indicates the rank of $e_i$ in the iteration over trees from
node $(A_i,o_i)$. (The first, highest-scoring tree has rank 1, the
second-highest-scoring tree has rank 2, etc.) The first tree is therefore
represented by the tuple $(1,1,\ldots,1)$. Whenever we report a tree
from node $N$, we prepare for the next iteration step by examining the tuple
$(n_1,\ldots,n_k)$ corresponding to the tree just reported, creating $k$ parse
trees corresponding to the tuples $(n_1+1,n_2,\ldots,n_k),
(n_1,n_2+1,\ldots,n_k),\ldots,(n_1,\ldots,n_{k-1},n_k+1)$ by requesting the
relevant subexpression trees from the child nodes of $N$, and pushing those $k$
trees into the priority queue of $N$.

These procedures for obtaining the best and next-best parse trees at OR- and
AND-nodes are similar to those we described previously \cite{ijdar2013}.
In that description, we required the implementation of the grammar
relations to satisfy certain properties in order for the tree exraction
algorithms to succeed. Here, similarly to the grammar definition, we
reformulate our assumptions more abstractly to promote independence between
theory and implementation.

The tree extraction algorithm is guaranteed to enumerate parse trees in strictly
decreasing order of any parse-tree scoring function $\score(e,o)$ (for $e$ an
expression and $o$ an observable) that satisfies the following assumptions.
In the assumptions, $S(\alpha,o)$ is a symbol recognition score for recognizing the
observable $o$ as the terminal symbol $\alpha$, $G(o)$ is a stroke grouping
score indicating the extent to which an observable $o$ is considered to be a
symbol, and $R_r((e_1,o_1),(e_2,o_2))$ is a relation score indicating the
extent to which the grammar relation $r$ is satisfied between the
interpretations $(e_1,o_1)$ and $(e_2,o_2)$. Larger values of these scoring
functions indicate higher recognition confidence.

\begin{enumerate}
	\item{The score of a terminal interpretation $(\alpha,t)$ is some
	function $g$ of the symbol and grouping scores,
	$$
		\score(\alpha,t) = g(S(\alpha,t), G(t)),
	$$
	and $g$ increases monotonically with $S(\alpha,o)$.
	}
	\item{The score of a nonterminal interpretation $(e,t) = (e_1 r \cdots r e_k,
	(t_1,\ldots,t_k))$ is some function $f_k$ of the sub-interpretation scores
	and the relation scores between adjacent sub-interpretations,
	\begin{align*}
		\score(e,t) = f_k( & \score(e_1,t_1),\ldots,\score(e_k,t_k),\\
		& R_r((e_1,t_1),(e_2,t_2)),\ldots,R_r((e_{k-1},t_{k-1}),(e_k,t_k))).
	\end{align*}
	and $f_k$ increases monotonically with all of its parameters.
	}
\end{enumerate}

These assumptions admit a number of reasonable
underlying scoring and combination functions. The next section will develop 
concrete scoring functions based on Bayesian probability theory and some
underlying low-level classification techniques.

\section{A Bayesian scoring function}
\label{sec:scoring-function}

At a high level, the scoring function $\score(e,o)$ of an interpretation of
some subset $o$ of the entire input $\hat{o}$ is organized as a Bayesian
network which is constructed specifically for $\hat{o}$.

\subsection{Model organization}
\label{sec:model-organization}

Given an input observable $\hat{o}$, the Bayesian scoring model
includes several families of variables:
\begin{itemize}
	\item{An expression variable $E_o \in \mathcal{L}(G) \cup
	\left\{ \nil \right\}$ for each $o \subseteq \hat{o}$, distributed over representable expressions.
	$E_o = \nil$ indicates that the subset $o$ has no meaningful interpretation.}
	\item{A symbol variable $S_o \in \Sigma \cup \left\{ \nil \right\}$ for each
	$o \subseteq \hat{o}$ indicating what terminal symbol $o$ represents.
	$S_o=\nil$ indicates that $o$ is not a symbol. ($o$ could be a larger
	subexpression, or could have $E_o=\nil$ as well.)}
	\item{A relation variable $R_{o_1,o_2} \in R \cup \left\{ \nil \right\}$
	indicating which grammar relation joins the subsets $o_1,o_2 \subset \hat{o}$.
	$R_{o_1,o_2} = \nil$ indicates that the subsets are not directly connected
	by a relation.}
	\item{A vector of grouping-oriented features $g_o$ for each $o \subseteq \hat{o}$.}
	\item{A vector of symbol-oriented features $s_o$ for each $o \subseteq \hat{o}$.}
	\item{A vector of relation-oriented features $f_o$ for each $o \subset \hat{o}$.}
	\item{A ``symbol-bag'' variable $B_o$ distributed over $\Sigma \cup \{ \nil \}$ for each $o
	\subseteq \hat{o}$.}
\end{itemize}

At any point in the parse tree extraction algorithm, a particular
interpretation $(e,o)$ of a subset of the input is under
consideration and must be scored. Each such interpretation corresponds to a
joint assignment to all of these variables, so our task is to define the joint
probability function 
$$
\Pr{\bigwedge_{o \subseteq \hat{o}} E_o, \bigwedge_{o \subseteq \hat{o}} S_o,
\bigwedge_{o_1,o_2 \subset \hat{o}} R_{o_1,o_2}, \bigwedge_{o \subseteq
\hat{o}} g_o, \bigwedge_{o \subseteq \hat{o}} s_o, \bigwedge_{o \subset
\hat{o}} f_o, \bigwedge_{o \subseteq \hat{o}} B_o}.
$$
in such a way that it may be calculated quickly.

The fact that joint variable assignments
arise from parse trees imposes severe constraints on which variables can
simultaneously be assigned non-$\nil$ values. For example, if
$E_o=\nil$, then $S_o=\nil$, and any $R_{o,\ast}$ or $R_{\ast,o}$ is also \nil.
Also, $E_o$ is a terminal expression iff $S_o \neq \nil$. In general, only
those variables representing meaningful subexpressions in the parse tree -- or
relationships between those subexpression -- will be non-$\nil$. In all valid
assignments, $B_o = S_o$ for all $o$; however the distribution of $S_o$ reflects recognition, while
that of $B_o$ reflects prior domain knowledge.

We take advantage of these constraints to achieve efficient calculation of the
joint probability. First, we factor the joint distribution as a Bayesian
network,
\begin{align*}
& \Pr{\bigwedge_{o \subseteq \hat{o}} E_o, \bigwedge_{o \subseteq \hat{o}} S_o,
\bigwedge_{o_1,o_2 \subset \hat{o}} R_{o_1,o_2}, \bigwedge_{o \subseteq
\hat{o}} g_o, \bigwedge_{o \subseteq \hat{o}} s_o, \bigwedge_{o \subset
\hat{o}} f_o, \bigwedge_{o \subset \hat{o}} B_o} \nonumber \\
= & \prod_{o' \subset \hat{o}} \CondPr{E_{o'}}{\bigwedge_{o \subset o'} E_o,
\bigwedge_{o \subseteq o'} S_o, \bigwedge_{o_1,o_2 \subset o'} R_{o_1,o_2}, \bigwedge_{o \subseteq
o'} g_o, \bigwedge_{o \subseteq o'} s_o, \bigwedge_{o \subset o'} f_o}
\nonumber \\
\times & \prod_{o \subseteq \hat{o}} \CondPr{S_o}{g_o,s_o} \nonumber \\
\times & \prod_{o_1,o_2 \subset \hat{o}}
\CondPr{R_{o_1,o_2}}{E_{o_1},E_{o_2},f_{o_1},f_{o_2}} \nonumber \\
\times & \prod_{o' \subseteq \hat{o}} \CondPr{B_{o'}}{\bigwedge_{o \subseteq
\hat{o}} S_{o}} \nonumber \\
\times & \prod_{o \subseteq \hat{o}} g_o \prod_{o \subseteq \hat{o}} s_o
\prod_{o \subset \hat{o}} f_o.
\end{align*}

Because of the number of variables, it is not feasible to compute this product
directly. Instead, we remove the feature variables as they are functions only
of the input, and factor out the product
$$
Z = \prod_{o \subseteq \hat{o}} \CondPr{S_o = \nil}{g_o,s_o} \times
\prod_{o_1,o_2 \subset \hat{o}} \CondPr{R_{o_1,o_2} =
\nil}{E_{o_1},E_{o_2},f_{o_1},f_{o_2}}.
$$
This division leaves
\begin{align*}
& \Pr{\bigwedge_{o \subseteq \hat{o}} E_o, \bigwedge_{o \subseteq \hat{o}} S_o,
\bigwedge_{o_1,o_2 \subset \hat{o}} R_{o_1,o_2}, \bigwedge_{o \subseteq
\hat{o}} g_o, \bigwedge_{o \subseteq \hat{o}} s_o, \bigwedge_{o \subset
\hat{o}} f_o, \bigwedge_{o \in \hat{O}} B_o} \nonumber \\
\propto & \prod_{o' \subset \hat{o},E_{o'}\neq\nil} \CondPr{E_{o'}}{\bigwedge_{o \subset o'} E_o,
\bigwedge_{o \subseteq o'} S_o, \bigwedge_{o_1,o_2 \subset o'} R_{o_1,o_2}, \bigwedge_{o \subseteq
o'} g_o, \bigwedge_{o \subseteq o'} s_o, \bigwedge_{o \subset o'} f_o} \nonumber \\
\times & \prod_{o \subseteq \hat{o},S_o \neq \nil}
\frac{\CondPr{S_o}{g_o,s_o}}{\CondPr{S_o=\nil}{g_o,s_o}} \nonumber \\
\times & \prod_{o_1,o_2 \subset \hat{o}, R_{o_1,o_2} \neq \nil}
\frac{\CondPr{R_{o_1,o_2}}{E_{o_1},E_{o_2},f_{o_1},f_{o_2}}}
{\CondPr{R_{o_1,o_2}=\nil}{E_{o_1},E_{o_2},f_{o_1},f_{o_2}}} \nonumber \\
\times & \prod_{o' \subset \hat{o}} \CondPr{B_{o'}}{\bigwedge_{o \subseteq
\hat{o}}S_{o'}}.
\end{align*}

Suppose our entire input observable contains $n$ strokes. Since the joint assignment to
these variables originates with an interpretation -- an expression $e$ along
with an observable $o$ -- at most
$n$ symbol variables may be non-$\nil$. An expression variable
$E_o$ is only non-$\nil$ when $o$ is a subset of the input corresponding to a
subexpression of $e$. A relation variable $R_{o_1,o_2}$ is only non-$\nil$ when
both $E_{o_1}$ and $E_{o_2}$
are non-$\nil$ and correspond to adjacent subexpressions of $e$. There are
thus $\mathcal{O}(n)$ non-$\nil$ variables in any joint assignment, so the
adjusted product may be calculated quickly. The RHS of the above expression
is the scoring function used during parse tree extraction.

To meaningfully compare the scores of different parse trees, the constant of proportionality
in this expression must be equal across all parse trees. Thus, the probabilities $\CondPr{S_o =
\nil}{g_o,s_o}$ and $\CondPr{R_{o_1,o_2} =
\nil}{E_{o_1},E_{o_2},f_{o_1},f_{o_2}}$ must be functions only of the input. To achieve this in
practice, we fix $E_{o_1} = E_{o_2} = \gen$ when computing $\CondPr{R_{o_1,o_2} =
\nil}{E_{o_1},E_{o_2},f_{o_1},f_{o_2}}$.

Dividing out $Z$ leaves four families of probability distributions which must
be defined. We will consider each of them in turn.

\subsection{Expression variables}
\label{sec:expression-variables}

The distribution of expression variables,
$$
E_{o'} \mid {\bigwedge_{o \subset o'} E_o,
\bigwedge_{o \subseteq o'} S_o, \bigwedge_{o_1,o_2 \subset o'} R_{o_1,o_2}, \bigwedge_{o \subseteq
o'} g_o, \bigwedge_{o \subseteq o'} s_o, \bigwedge_{o \subset o'} f_o},
$$
is deterministic. In a valid joint
assignment (i.e., one derived from a parse tree), each non-$\nil$ conditional
expression variable $E_o$ is a subexpression of $E_{o'}$, and the non-$\nil$
relation variables indicate how these subexpressions are joined together.
This information completely determines what expression $E_{o'}$ must
be, so that expression is assigned probability 1.

\subsection{Symbol variables}
\label{sec:symbol-variables}

The distribution of symbol variables, ${S_o} \mid {g_o,s_o}$, is based on the
stroke grouping score $G(o)$ and the symbol recognition scores $S(\alpha,o)$
(These scores are described below.)

The $\nil$ probability of $S_o$ is allocated as
$$
\CondPr{S_o = \nil}{g_o,s_o} = 1 - \frac{N}{N+1},
$$
where $N = \log ( 1 + G(o) \max_\alpha \{ S(\alpha,o) \} )$, and the remaining
probability mass is distributed proportionally to $S(o,\alpha)$:
$$
\CondPr{S_o = \alpha}{g_o,s_o} \propto \frac{S(\alpha,o)}{\sum_{\beta \in
\Sigma} S(\beta,o)}.
$$

\subsubsection{Stroke grouping score}
\label{sec:stroke-grouping-score}

The stroke grouping function $G(o)$ is based on intuitions about what
characteristics of a group of strokes indicate that the strokes belong to the same
symbol. These intuitions may be summarized by the following logical predicate
using the variables defined in Table \ref{tab:grouping-variables}:
$$
G = \left(D_{in} \vee \left( L_{in} \wedge \neg C_{in} \right) \right) \wedge
\left( \neg L_{out} \vee C_{out} \right).
$$

\begin{table}[htp]
\centering
\small
\begin{tabular}{c|l}
Name & Meaning \\
\hline
$G$ & A group exists \\
$D_{in}$ & Small distance between strokes within group \\
$L_{in}$ & Large overlap of strokes within group \\
$C_{in}$ & Containment notation used within group \\
$L_{out}$ & Large overlap of group and non-group strokes \\
$C_{out}$ & Containment notation used in group or overlapping non-group strokes
\end{tabular}
\caption{\label{tab:grouping-variables}Variables names for grouping predicate.}
\end{table}

Using DeMorgan's law, this predicate may be re-written as
\begin{align*}
G & = \neg \left( \neg D_{in} \wedge \neg \left( L_{in} \wedge \neg C_{in} \right) \right) \wedge \neg \left( L_{out} \wedge \neg C_{out} \right) \\
& = \neg \left( \neg D_{in} \wedge \neg X_{in} \right) \wedge \neg X_{out},
\end{align*}
where $X_{\ast} = L_{\ast} \wedge \neg C_{\ast}$.

To make this approach quantitative, we measure several aspects of $o$, one for
each of the RHS variables defined above:
\begin{itemize}
	\item{$d = \min \left\{ \mbox{dist}(s',s) : s' \in g \right\}$, where $\mbox{dist}(s_1,s_2)$
	is the minimal distance between the curves traced by $s_1$ and $s_2$;}
	\item{$\ell_{in} = \overlap(g,s)$, where $\overlap(g,s)$ is the area
	of the intersection of the bounding boxes of $g$ and $s$ divided by the
	area of the smaller of the two boxes;}
	\item{$c_{in} = \max \left( C(s), \max \left\{ C\left(s' \right) : s' \in g \right\}
	\right)$, where $C\left(s \right)$ is the extent to which the stroke $s$ resembles
	a containment notation, as described below;}
	\item{$\ell_{out} = \max \left\{ \overlap(g \cup \{s\}, s') : s' \in S
	\setminus \left(g \cup \left\{ s \right\} \right) \right\}$;}
	\item{$c_{out} = \max \left( c_{in}, C\left( s' \right) \right)$, where $s'$ is the
	maximizer for $\ell_{out}$.}
\end{itemize}

Together, these five measurements form the grouping-oriented feature vector $g_o$.
The grouping function $G(o)$ is defined as a conditional probability:
\begin{align*}
G(o) & = \CondPr{G}{d,\ell_{in},c_{in},\ell_{out},c_{out}} \\
& = \left( 1 - \CondPr{\neg D_{in}}{d} \CondPr{\neg X_{in}}{\ell_{in},c_{in}} \right)^\beta
\CondPr{\neg X_{out}}{\ell_{out},c_{in},c_{out}}^{1-\beta}
\end{align*}
where we set
\begin{align*}
\label{eqn:grouping-parameter-functions}
\CondPr{\neg D_{in}}{d} & = \left( 1 - e^{-d / \lambda} \right)^{\alpha} \\
\CondPr{\neg X_{in}}{\ell_{in}, c_{in}} & = \left( 1 - \ell_{in} \left( 1 - c_{in} \right) \right)^{\left(1-\alpha\right)} \\
\CondPr{\neg X_{out}}{\ell_{out}, c_{out}, c_{in}} & = 1 - \ell_{out}
\left( 1 - \max\left(c_{in}, c_{out}\right) \right).
\end{align*}
This definition is loosely based on treating the boolean variables from our
logical predicate as random and independent. $\lambda$ is estimated from training
data, while $\alpha,\beta$ were both fixed via experiments at $9/10$.

To account for containment notations in the calculation of $c_{in}$ and
$c_{out}$, we annotate each symbol supported by the symbol recognizer with
a flag indicating whether it is used to contain other symbols. (Currently only
the $\sqrt{\phantom{x}}$ symbol has this flag set.) Then the symbol recognizer
is invoked to measure how closely a stroke or group of strokes resembles a
container symbol.

\subsubsection{Symbol recognition score}
\label{sec:symbol-recognition-score}

The symbol recognition score $S(\alpha,o)$ for recognizing an input subset $o$
as a terminal symbol $\alpha$ uses a combination of four distance-based
matching techniques within a template-matching scheme. In this scheme, the
input strokes in $o$ are matched against \textit{model} strokes taken from a stored
training example of $\alpha$, yielding a match distance. This
process is repeated for each of the training examples of $\alpha$, and the two
smallest match distances are averaged for each technique, giving four minimal
distances $d_1,d_2,d_3,d_4$. The distances are then combined in a weighted
sum
$$
s_\alpha = \sum_{i=1}^4 w_i Q_i(d_i),
$$
where $Q_i : \mathbb{R} \rightarrow [0,1]$ is the quantile function for the
distribution of values emitted by the $i$th technique. These quantiles are
approximated by piecewise linear functions and serve to normalize
the outputs of the various distance functions so that they may be meaningfully
arithmetically combined. The weights are optimized from training data.
Finally, $S(\alpha,o) = s_\alpha^{-2}$ so that small distances correspond to
large scores.  The symbol recognition feature vector $s_o$ in the Bayesian
model contains the values $S(\alpha,o)$ for all terminals $\alpha$.

The four distance-based techniques used are as follows:
\begin{itemize}
	\item{A fast elastic matching variant \cite{das10} which finds a matching
	between the points of the input and model strokes and minimizes the sum
	of distances between matched points.}
	\item{A functional approximation technique \cite{golubitsky09} which models
	stroke coordinates as parametric functions expressed as truncated
	Legendre-Sobolev series. The match distance is the 2-norm between coefficient vectors.}
	\item{An offline technique \cite{thammano06} based on rasterizing strokes,
	treating filled pixels as points in Euclidean space, and computing the Hausdorff distance
	between those point sets.}
	\item{A feature-vector method using the 2-norm between vectors containing
	the following normalized measurements of the input and model strokes: bounding box
	position and size, first and last point coordinates, and total arclength.}
\end{itemize}

\subsection{Relation variables}
\label{sec:relation-variables}

The distribution of relation variables, $R_{o_1,o_2} \mid
{E_{o_1},E_{o_2},f_{o_1},f_{o_2}}$, is similar to the symbol variable case in
that it is based on the output of a lower-level recognition
method, with some of the probability mass being reserved for the $\nil$
assignment. In particular, let $R(r)$ denote the extent to
which the observables $o_1$ and $o_2$ appear to satisfy the grammar relation
$r$. Then we put
$$
\CondPr{R_{o_1,o_2}=\nil}{E_{o_1},E_{o_2},f_{o_1},f_{o_2}} = 1 - \frac{N}{N+1},
$$
where $N = \log ( 1 + \max_r R(r) )$, and distribute the remaining probability
mass proportionally to $R(r)$:
$$
\CondPr{R_{o_1,o_2}=r}{E_{o_1},E_{o_2},f_{o_1},f_{o_2}} \propto
\frac{R(r)}{\sum_{u \in R} R(u)}.
$$

$R(r)$ is determined using two sources of information. The
first is geometric information about the bounding boxes of $o_1$ and $o_2$. Let
$\ell(o), r(o), t(o),$ and $b(o)$ respectively denote the left-, right-, top-,
and bottom-most coordinates of an observable $o$. These four values form the
relation-oriented feature vector $f_o$ in the Bayesian model. From two such
vectors $f_{o_1}$ and $f_{o_2}$, the following relation classification
features are obtained ($N$ is a normalization factor based on the sizes of
$o_1$ and $o_2$):
\begin{align*}
	f_1 & = \frac{\ell(o_2) - \ell(S_1)}{N} \\
	f_2 & = \frac{r(o_2) - r(o_1)}{N} \\
	f_3 & = \frac{\ell(o_2) - r(o_1)}{N} \\
	f_4 & = \frac{b(o_2) - b(o_1)}{N} \\
	f_5 & = \frac{t(o_2) - t(o_1)}{N} \\
	f_6 & = \frac{t(o_2) - b(o_1)}{N} \\
	f_7 & = \overlap(o_1, o_2)
\end{align*}

The second type of information is prior knowledge about how two types of expressions
are arranged when placed in each grammar relation. To represent this, each
expression $e$
is associated with one or more \textit{relational classes} by a function
$\class(e)$. Each terminal symbol is itself a relational class (e.g., $\int,
x, +$). Several more classes represent symbol stereotypes, as
summarized in Table \ref{tab:relation-class-labels}. Finally, there are three
generic classes: $\sym$ (any symbol), $\expr$ (any multi-symbol expression),
and $\gen$ (for ``generic''; any expression whatsoever).

\begin{table}[htp]
\centering
\small
\begin{tabular}{|c|c|}
\hline
Class name & Example symbols \\
\hline
Baseline & a c x \\
Ascender & 2 A h \\
Descender & g y \\
Extender & ( ) \\
Centered & + = \\
i & i \\
j & j \\
Large-Extender & $\int$ \\
Root & $\sqrt{\phantom{x}}$ \\
Horizontal & - \\
Punctuation & . , ' \\
\hline
\end{tabular}
\caption{\label{tab:relation-class-labels}Relational class labels.}
\end{table}

These labels are hierarchical with respect to specificity. For example, 
the symbol $c$ is also a baseline symbol, any baseline
symbol is also a $\sym$ expression, and any $\sym$ expression is also a $\gen$
expression.

We construct a classed scoring function $R_{c_1,c_2}$ for every pair
$c_1,c_2$ of classes. Then, to evaluate $R(r)$, we first try the most specific
relevant classed scoring function. If, due to a lack of training data, it is
unable to report a result, we fall back to a less-specific classed function.
This process of gradually reducing the specificity of class labels continues
until a score is obtained or both labels are the generic label $\gen$.

Each of the classed scoring functions is a naive Bayesian model that 
treats each of the features $f_i$ as drawn from independent normally-distributed
variables $F_i$. So
$$
R_{c_1,c_2}(r) = \Pr{R=r} \prod_{i=1}^7 \CondPr{F_i = f_i}{R = r},
$$
where $R$ is a categorical random variable distributed over the grammar
relations. The parameters of these distributions are estimated from training
data.

When calculating a product of the form above, we check whether the true
mean of each distribution lies within a 95\% confidence interval of the
training data's sample mean. If not, then the classed scoring function
$R_{c_1,c_2}(r)$ reports failure, and we fall back to less-specific relational
class labels as described above.

\subsection{``Symbol-bag'' variables}
\label{sec:symbol-bag-variables}

The role of the symbol-bag variables $B_o$ is to apply a prior distribution to
the terminal symbols appearing in the input, and to adapt that distribution
based on which symbols are currently present. Because those symbols are not
known with certainty, such adaptation may only be done imperfectly.

We collected symbol occurence and co-occurence rates from the Infty project
corpus \cite{suzuki05} as well as from the \LaTeX\, sources of University of
Waterloo course notes for introductory algebra and calculus courses. The
symbol occurence rate $r(\alpha)$ is the number of expressions in which the
symbol $\alpha$ appeared and the co-occurence rate $c(\alpha,\beta)$ is the number
of expressions containing both $\alpha$ and $\beta$.

In theory, each $B_o$ variable is identically distributed and satisfies
$$
\CondPr{B_o=\nil}{S_o=\nil,\bigwedge_{o' \subseteq \hat{o}} S_{o'}} = 1
$$
and
$$
\CondPr{B_o=\alpha}{S_o=\delta,\bigwedge_{o' \subseteq \hat{o}} S_{o'}} \propto r(\alpha) + \sum_{\beta \in \Sigma} \Pr{S_{o'} = \beta
\text{ for some } o' \subseteq \hat{o}} \frac{c(\alpha,\beta)}{r(\beta)},
$$
for any non-$\nil$ $\delta \in \Sigma$ (recall that $\hat{o}$ is the entire
input observable).

The trivial $\nil$ case allows us to remove all terms with $B_{o'}=\nil$ from
the product $$\prod_{o' \subset \hat{o}} \CondPr{B_{o'}}{\bigwedge_{o \subseteq
\hat{o}} S_{o}}$$ in our
scoring function.

In the non-$\nil$ case, note that for any subset $o'$ of the input for which
the grouping score $G(o')$ is 0, we have $S_{o'} = \nil$ with probability one;
thus in the $\Pr{S_{o'} = \beta \text{ for some } o' \subseteq o}$ term, we
need only consider those subsets of $o$ with a non-zero grouping score. Rather than
assuming independence and iterating over all such subsets, or developing a more
complex model, we simply approximate this probability by
$$\max_{o' \subseteq o, G(o') > 0} \CondPr{S_{o'} = \beta}{g_{o'},s_{o'}}.$$

Evaluation of these symbol-bag probabilities proceeds as follows. Starting with an
empty input, only the $r(\alpha)$ term is used. The distribution is updated
incrementally each time symbol recognition is performed on a candidate stroke group.
The set of candidate groups is updated each time a new stroke is
drawn by the user. When a group corresponding to an observable $o$ is
identified, the symbol recognition and stroke grouping processes induce the
distribution of $S_o$, which we use to update the distribution of the $B_\ast$
variables by checking whether any of the $\max$ expressions need to be revised.
Intuitively, $B_\ast$ treats the ``next'' symbol to be recognized as being
drawn from a bag full of symbols. If a previously-unseen symbol $\beta$ is drawn,
then more symbols are added to the bag based on the co-occurence counts of $\beta$.

This process is similar to updating the parameter vector of a Dirichlet
distribution, except that the number of times a given symbol appears within an
expression is irrelevant. We are concerned only with how likely it is that the
symbol appears at all.

\section{Accuracy evaluation}
\label{sec:evaluation}

We performed two accuracy evaluations of the MathBrush recognizer.
The first replicates the 2011 and 2012 CROHME evaluations
\cite{crohme2011,crohme2012}, which focus on the top-ranked expression only.
To evaluate the efficacy of our recognizer's correction mechanism, we
also replicated a user-focused evaluation which was initially developed to
test an earlier version of the recognizer \cite{ijdar2013}.

\subsection{CROHME evaluation}
\label{sec:crohme-evaluation}

Since 2011, the CROHME math recognition competition has invited researchers to
submit recognition systems for comparative accuracy evaluation. In previous
work, we compared the accuracy of a version of the MathBrush
recognizer against that of the 2011 entrants \cite{ijdar2013}. We also participated
in the 2012 competition, placing second behind a corporate entrant
\cite{crohme2012}. In this
section we will evaluate the current MathBrush recognizer described by this
paper against its previous versions as well as the other CROHME entrants.

The 2011 CROHME data is divided into two parts, each of which includes training and
testing data. The first part includes a relatively small selection of
mathematical notation, while the second includes a larger selection. For 
details, refer to the competition paper \cite{crohme2011}. The 2012 data is
similarly divided, but also includes a third part, which we omit from this
evaluation as it included notations not used by MathBrush.

For each part of this evaluation, we trained our symbol recognizer on
the same base data set as in the previous evaluation and augmented that
training data with all of the symbols
appearing in the appropriate part of the CROHME training data. The relation
and grouping systems were trained on the 2011 Waterloo corpus. We used the
grammars provided in the competition documentation.

The accuracy of our recognizer was measured using a perl script provided by the
competition organizers. In this evaluation, only the top-ranked parse
was considered. There are four accuracy measurements: stroke reco.
indicates the percentage of input strokes which were correctly recognized and
placed in the parse tree; symbol seg. indicates the percentage of symbols for
which the correct strokes were properly grouped together; symbol reco.
indicates the percentage of symbol recognized correctly, out of those
correctly grouped; finally, expression reco. indicates the percentage of
expressions for which the top-ranked parse tree was exactly correct.

In these tables, ``University of Waterloo (prob.)'' refers to the recognizer
described in this paper, ``University of Waterloo (2012)'' refers to the version submitted
to the CROHME 2012 competition, and ``University of Waterloo (2011)'' refers to the
version used for a previous publication \cite{ijdar2013}. Both of these
previous versions used a fuzzy set formalism to capture and manage ambiguous
interpretations. The remaining rows
are reproduced from the CROHME reports \cite{crohme2011,crohme2012}.

\begin{table}[htp]
\centering
\small
\begin{tabular}{|l|c|c|c|c|}
\hline
\multicolumn{5}{|c|}{Part 1 of corpus} \\
\hline
Recognizer & Stroke reco. & Symbol seg. & Symbol reco. & Expression reco. \\
\hline
\textbf{University of Waterloo (prob.)} & \textbf{93.16} & \textbf{97.64} & \textbf{95.85} & \textbf{67.40} \\
University of Waterloo (2012) & 88.13 & 96.10 & 92.18 & 57.46 \\
University of Waterloo (2011) & 71.73 & 84.09 & 85.99 & 32.04 \\
Rochester Institute of Technology & 55.91 & 60.01 & 87.24 & 7.18 \\
Sabanci University & 20.90 & 26.66 & 81.22 & 1.66 \\
University of Valencia & 78.73 & 88.07 & 92.22 & 29.28 \\
Athena Research Center & 48.26 & 67.75 & 86.30 & 0.00 \\
University of Nantes & 78.57 & 87.56 & 91.67 & 40.88 \\
\hline
\multicolumn{5}{|c|}{Part 2 of corpus} \\
\hline
Recognizer & Stroke reco. & Symbol seg. & Symbol reco. & Expression reco. \\
\hline
\textbf{University of Waterloo (prob.)} & \textbf{91.64} & \textbf{97.08} & \textbf{95.53} & \textbf{55.75} \\
University of Waterloo (2012) & 87.08 & 95.47 & 92.20 & 47.41 \\
University of Waterloo (2011) & 66.82 & 80.26 & 86.11 & 20.11 \\
Rochester Institute of Technology & 51.58 & 56.50 & 91.29 & 2.59 \\
Sabanci University & 19.36 & 24.42 & 84.45 & 1.15 \\
University of Valencia & 78.38 & 87.82 & 92.56 & 19.83 \\
Athena Research Center & 52.28 & 78.77 & 78.67 & 0.00 \\
University of Nantes & 70.79 & 84.23 & 87.16 & 22.41 \\
\hline
\end{tabular}
\caption{\label{tab:crohme2011}Evaluation on CROHME 2011 corpus.}
\end{table}

\begin{table}[htp]
\centering
\small
\begin{tabular}{|l|c|c|c|c|}
\hline
\multicolumn{5}{|c|}{Part 1 of corpus} \\
\hline
Recognizer & Stroke reco. & Symbol seg. & Symbol reco. & Expression reco. \\
\hline
\textbf{University of Waterloo (prob.)} & \textbf{91.54} & \textbf{96.56} & \textbf{94.83} & \textbf{66.36} \\
University of Waterloo (2012) & 89.00 & 97.39 & 91.72 & 51.85 \\
University of Valencia & 80.74 & 90.74 & 89.20 & 35.19 \\
Athena Research Center & 59.14 & 73.31 & 79.79 & 8.33 \\
University of Nantes & 90.05 & 94.44 & 95.96 & 57.41 \\
Rochester Institute of Technology & 78.24 & 92.81 & 86.62 & 28.70 \\
Sabanci University & 61.33 & 72.11 & 87.76 & 22.22 \\
Vision Objects & 97.01 & 99.24 & 97.80 & 81.48 \\
\hline
\multicolumn{5}{|c|}{Part 2 of corpus} \\
\hline
Recognizer & Stroke reco. & Symbol seg. & Symbol reco. & Expression reco. \\
\hline
\textbf{University of Waterloo (prob.)} & \textbf{92.74} & \textbf{95.57} & \textbf{97.05} & \textbf{55.18} \\
University of Waterloo (2012) & 90.71 & 96.67 & 94.57 & 49.17 \\
University of Valencia & 85.05 & 90.66 & 91.75 & 33.89 \\
Athena Research Center & 58.53 & 72.19 & 86.95 & 6.64 \\
University of Nantes & 82.28 & 88.51 & 94.43 & 38.87 \\
Rochester Institute of Technology & 76.07 & 89.29 & 91.21 & 14.29 \\
Sabanci University & 49.06 & 61.09 & 88.36 & 7.97 \\
Vision Objects & 96.85 & 98.71 & 98.06 & 75.08 \\
\hline
\end{tabular}
\caption{\label{tab:crohme2012}Evaluation on CROHME 2012 corpus.}
\end{table}

These results indicate that the MathBrush recognizer is significantly more
accurate than all other CROHME entrants except for Vision Objects recognizer, which
is much more accurate again. Moreover, the performance of the current version of the MathBrush
recognizer is a significant improvement over previous versions, except in the
case of symbol segmentation (stroke grouping) on part 2 of the 2012 data set. This is
possibly because the MathBrush 2012 recognizer submitted to CROHME
was fairly well-tuned to the 2012 training data, whereas we did not adjust any
recognition parameters for the evaluation presented in this paper.

With respect to the Vision Objects recognizer, a short description may be
found in the CROHME reports \cite{crohme2012,crohme2013}. Many similarities
exist between their methods and our own: notably, a systematic combination of
the stroke grouping, symbol recognition, and relation classifiaction problems,
relational grammars in which each production is associated with a single
geometric relationship, and a combination of online and offline methods for
symbol recognition. The main significant difference between our approaches
appears to be the statistical language model of Vision Objects and their
use of very large private training corpora (hundreds of thousands for the
language model, with 30000 used for training the CROHME 2013 system -- the
precise distinction between these training sets is unclear).

\subsection{Evaluation on Waterloo corpus}
\label{sec:waterloo-evaluation}

In previous work, we proposed a user-oriented accuracy metric that measured
how accurate an expression was recognized by counting how many corrections
needed to be made to the results \cite{ijdar2013}. This metric is similar to a
tree-based edit distance in that each edit operation corresponds to selecting
one of the alternative recognition results offered by the recognizer on a
particular subset of the input.

To facilitate meaningful comparison with our previous results, we have replicated
the same evaluation as closely as possible. The evaluation consists of two
scenarios. In the first (``\textit{default}'') scenario, the 1536 corpus
transcriptions containing fewer than four symbols were used as a training set,
with the remainder of the 3674-transcription corpus used as a testing set.
Each symbol in the training set was extracted and added to the library of
training examples. This library also contained roughly 5-15 samples of each symbol
collected from writers not appearing in the corpus, which ensured that all
symbols possessed an adequate number of training examples. Because the training
transcriptions contained only a few symbols each, they did not yield sufficient
data to adequately train the relation classifier. We therefore augmented the
relation classifier training data with larger transcriptions from a 2011
collection study \cite{sbim11}. (The symbol recognizer was not trained on this
data.)

The second (``\textit{perfect}'') scenario disregarded stroke grouping and
symbol recognition results, which are the largest sources of ambiguity in our
recognizer, and evaluated the quality of expression parsing and relation
classification. The training and testing sets for this scenario were identical
to those for the default scenario, but terminal symbol identities were extracted
directly from transcription ground truth prior to recognition, bypassing the stroke
grouping and symbol recognition systems altogether.

Following recognition, each transcription is assigned to one of the following
classes:

\begin{enumerate}
\item{\textit{Correct}: No corrections were required. The top-ranked
interpretation was correct.}
\item{\textit{Attainable}: The correct interpretation was obtained from the
recognizer after one or more corrections.}
\item{\textit{Incorrect}: The correct interpretation could not be obtained from the
recognizer.}
\end{enumerate}

In the original experiment, the ``incorrect'' class was divided into both
``incorrect'' and ``infeasible''. The infeasible class counted
transcriptions for which the correct symbols were not identified by the symbol
recognizer, making it easier to distinguish between symbol recognition
failures and relation classification failures. In the current version of the
system, many symbols are recognized through a combination of symbol and
relation classification, so the distinction between feasible and infeasible is
no longer useful. We have therefore merged the infeasible and incorrect
classes for this experiment. Table \ref{tab:evaluation-waterloo} summarizes
the recognizer's accuracy and the average number of symbol and structural
corrections required in each of these scenarios.

\begin{table}[htp]
\small
\centering
\begin{tabular}{|l|c|c|c|c|c|}
\hline
Recognizer & \% Correct & \% Attainable & \% Incorrect & \# Sym. Corr. & \# Struct. Corr. \\
\hline
\multicolumn{6}{|c|}{Default Scenario} \\
\hline
Uni. of Waterloo (prob.) & 33.26 & 46.68 & 20.06 & 0.53 & 0.33 \\
Uni. of Waterloo (2011) & 18.7 & 48.83 & 32.47 & 0.67 & 0.58 \\
\hline
\multicolumn{6}{|c|}{Perfect Scenario} \\
\hline
Uni. of Waterloo (prob.) & 85.22 & 10.71 & 4.07 & n/a & 0.11 \\
Uni. of Waterloo (2011) & 82.68 & 15.68 & 1.64 & n/a & 0.17 \\
\hline
\end{tabular}
\caption{\label{tab:evaluation-waterloo}Accuracy rates and correction counts
from evaluation on the Waterloo corpora.}
\end{table}

In the default scenario, the rates of attainability (i.e., the sum of correct and
attainable columns) and correctness were both significantly higher for the recognizer
described in this paper than for our earlier system, which was based on fuzzy sets. The
average number of corrections required to obtain the correct interpretation was also
lower for the new recognizer.

In the perfect scenario, the rates were much closer. The probabilistic system achieved a
somewhat higher correctness rate, but in this scenario had a slightly lower
attainability rate than the fuzzy variant. Rather than this indicating a problem with the
probabilistic recognizer, it likely indicates that the hand-coded rules of the
fuzzy relation classifier were too highly tuned for the 2009 training data.
Further evidence of this ``over-training'' of the 2011 system is pointed out by
MacLean \cite{phd-thesis}.

\section{Conclusions and future work}
\label{sec:conclusions}

As shown in the evaluation above, the MathBrush recognizer is competitive
with other state-of-the-art academic recognition systems, and its accuracy is
improving over time as work continues on the project. At a high level, the
recognizer organizes its input via relational context-free grammars, which
represent the concept of recognition quite naturally via interpretations --
a grammar-derivable expression paired with an observable set of input strokes.

By introducing some constraints on how inputs may be partitioned, we derived
an efficient parsing algorithm derived from Unger's method. The output of this
algorithm is a parse forest, which simultaneously represents all recognizable
parses of the input in a fairly compact data structure. From the parse forest,
individual parse trees may be obtained by the extraction algorithm described
in Section \ref{sec:tree-extraction}, subject to some abstract restrictions on
how parse trees are scored.

The tree scoring function is the primary novel contribution of this paper. We
construct a Bayesian network based on the input and use the fact that variable
assignments are derived from parse trees to eliminate the majority of
variables from probability calculations. Three of the five remaining families
of distributions -- stroke grouping, symbol identity, and relation identity -- are
based on the results of lower-level classification systems.
The remaining distribution families (expressions and symbol priors) are
respectively derived from the grammar structure and training data.

The Bayesian network approach provides a systematic and extensible foundation
for scoring parse trees. It is straightforward to combine underlying scoring
systems in a well-specified way, and it is easy to modify those underlying
systems without disrupting the model as a whole. However, some improvements
are still desirable.

In particular, we wish to allow score interactions between sibling nodes in a
parse tree, not just between parents and children. This would, for example,
let $\cos \omega$ be assigned a higher score than $\cos w$ even if $w$ had a
higher symbol recognition score than $\omega$, provided that the combination
of $\cos$ and $\omega$ was more prevalent in training data than the
combination of $\cos$ and $w$. There are a few ways in which such information
may be included in the Bayesian model, although we currently lack a sufficiently
large set of training expressions to effectively train the result.

More importantly, such a change to the Bayesian model would also necessitate
an adjustment of our parsing methods, since the example just mentioned
violates the restrictions on scoring functions required by the parse
tree extraction algorithm. How to permit more flexibility in subexpression
combinations while maintaining reasonable speed and keeping track of multiple
interpretations is our main focus for future research.



\bibliography{pen-math}

\end{document}